\documentclass{article}
\usepackage{arxiv}
\usepackage[utf8]{inputenc} % allow utf-8 input
\usepackage[T1]{fontenc}    % use 8-bit T1 fonts
\usepackage{hyperref}       % hyperlinks
\usepackage{url}            % simple URL typesetting
\usepackage{booktabs}       % professional-quality tables
\usepackage{amsfonts}       % blackboard math symbols
\usepackage{nicefrac}       % compact symbols for 1/2, etc.
\usepackage{microtype}      % microtypography
\usepackage{lipsum}		% Can be removed after putting your text content
\usepackage{graphicx}
\usepackage{natbib}
\usepackage{doi}
\usepackage{util}
\usepackage{multirow}

\title{Bridging the Projection Gap: Overcoming Projection Bias Through Parameterized Distance Learning}

\author{Chong Zhang \\
	Xi'an Jiaotong-Liverpool University\\
	\And
	Mingyu Jin \\
        Northwestern University \\
        \And
        Qinkai Yu \\
        University of Liverpool \\
        \And
        Haochen Xue \\
        Xi'an Jiaotong-Liverpool University\\
        \And
        Shreyank N Gowda \\
        University of Oxford \\
        \And
        Xiaobo Jin\thanks{Corresponding Author. Xiaobo Jin: Xiaobo.Jin@xjtlu.edu.cn} \\
        Xi'an Jiaotong-Liverpool University\\
}

\renewcommand{\shorttitle}{}

\begin{document}
\maketitle

\begin{abstract}

Generalized zero-shot learning (GZSL) aims to recognize samples from both seen and unseen classes using only seen class samples for training. However, GZSL methods are prone to bias towards seen classes during inference due to the projection function being learned from seen classes. Most methods focus on learning an accurate projection, but bias in the projection is inevitable. We address this projection bias by proposing to learn a parameterized Mahalanobis distance metric for robust inference. Our key insight is that the distance computation during inference is critical, even with a biased projection. We make two main contributions - (1) We extend the VAEGAN (Variational Autoencoder \& Generative Adversarial Networks) architecture with two branches to separately output the projection of samples from seen and unseen classes, enabling more robust distance learning.  (2) We introduce a novel loss function to optimize the Mahalanobis distance representation and reduce projection bias. Extensive experiments on four datasets show that our approach outperforms state-of-the-art GZSL techniques with improvements of up to 3.5 \% on the harmonic mean metric.

\end{abstract}

\section{Introduction}

Deep learning (DL) models have achieved recent advances in computer vision and have gained widespread popularity due to their ability to provide end-to-end solutions from feature extraction to classification. Despite their success, traditional deep learning models require extensive labelled data for each category. However, collecting large-scale datasets is a challenging problem due to the time and expenses related to it. Zero-shot learning (ZSL) \cite{palatucci2009zero, larochelle2008zero} technology provides a good solution to this challenge. ZSL aims to train a model that can classify images and realize knowledge transfer from seen classes (source classes) to unseen classes (target domain) through semantic information, which is leveraged to bridge the gap between seen and unseen classes. 
\begin{figure*}[!ht]
    \centering
    \includegraphics[width=0.8\textwidth]{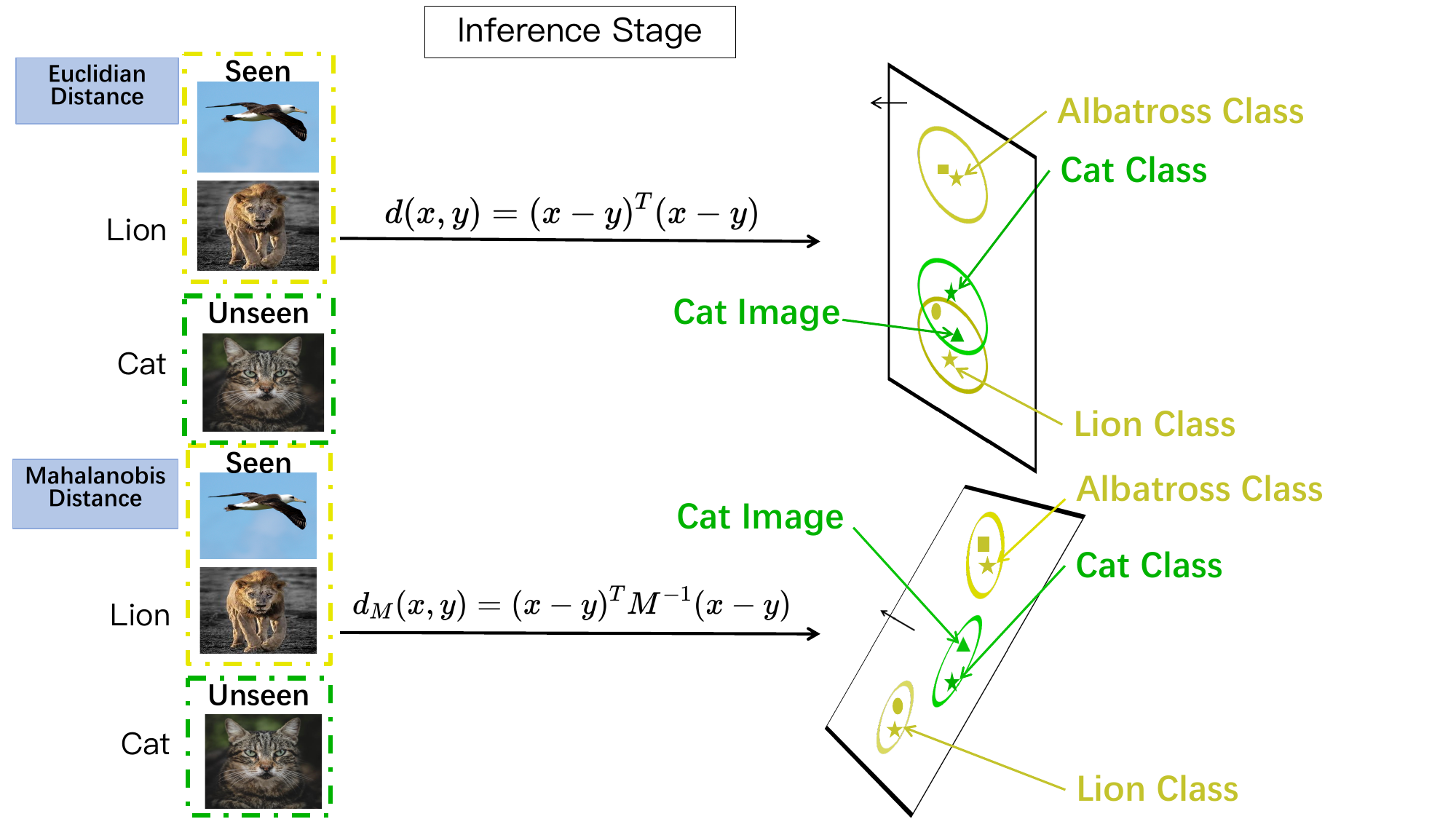}
    \caption{Demonstration of how the Mahalanobis distance compensates for the biased nature of GZSL in the projection space: when image instances and class descriptions are biased in the projection space, an image from the Cat class (indicated by the green triangle) will be misclassified into the Lion class (deep yellow pentagram) according to the Euclidean distance (the top part); however, the image will be correctly classified into the Cat class according to the Mahalanobis distance (the bottom part).}
    \label{fig:overview}
\end{figure*}
In real-world scenarios, data samples from seen classes often outnumber those from unseen classes. The generalized zero-shot learning (GZSL) paradigm addresses this, aiming to classify both seen and unseen class samples concurrently. The crux of most GZSL techniques is to determine an embedding or projection function that links seen class visual features to their respective semantic vectors. This function then aids in classifying test samples based on their proximity to unseen class semantic vectors. In doing so, GZSL methods bridge the gap between the high-dimensional visual space and the semantic attribute space, facilitating the transfer of learned knowledge to accurately classify unseen classes by exploiting their semantic relationships.

Most GZSL methods learn embedding/projection functions to associate seen low-level visual feature classes with their corresponding semantic vectors. The learned function is used to compute the distance between the semantic representation of the class and the projected representation of the sample and classify them to the nearest class. Since each entry of an attribute vector represents a description of that class, class descriptions with similar features are expected to contain similar attribute vectors in the semantic space. However, in visual space, classes with similar properties can be quite different. Therefore, finding a precise and suitable embedding space is a challenging task. Otherwise, it may lead to an ambiguity problem with visual semantics.

There is a problem of projection domain bias in the embedding methods used for the Generalized Zero-Shot Learning (GZSL) task. On one hand, vision and semantics are located in two distinct spaces; on the other hand, the samples of seen and unseen classes do not intersect, leading to potential differences in their distributions. Consequently, without appropriate adjustments to the embedding space for unseen classes, a problem of projection domain bias may arise \cite{fu2015transductive, zhao2017zero, jia2019deep}. Since GZSL methods must recognize both seen and unseen categories during inference but only have access to visual features of seen categories during training, they are usually biased towards seen categories. To address this issue, inductive methods incorporate additional constraints or information about the seen classes, whereas transductive methods leverage available information to mitigate the problem of projection domain shifts \cite{cheraghian2020transductive, rahman2019transductive, guan2019extreme, huo2018zero}.

Typical GZSL algorithms perform the following three steps during the inference process: 1) Project the image into the space where the class semantic vector is located; 2) Calculate the Euclidean distance between the projection vector and each semantic vector; 3) Classify the image to the class with the closest distance. We argue that Mahalanobis distance considers the correlation of features, unlike Euclidean distance. This is critical for addressing projection bias, especially when the class distribution changes significantly. Therefore, in our work, we will learn the Mahalanobis distance to influence the decision in step 3 to mitigate the impact of the projection bias in step 1.

Building on this foundation, we explore the integration of the Mahalanobis distance within the VAEGAN framework \cite{xian2019f}, which synergizes the strengths of VAE and WGAN models \cite{xian2018feature} for sampling from data distributions. Our extension introduces a dual-branch structure within the VAEGAN model, specifically designed to accommodate the learning of Mahalanobis distance. The upper branch generates unseen class images from the seen class via a generative network, simulating the inference stage for classifying images of unseen classes. Concurrently, the lower branch focuses on directly learning projective representations for seen class images. To optimize this architecture, we propose a novel loss function based on Mahalanobis distance, aimed at minimizing the distance for projections within the same branch while maximizing the separation for projections across different branches.

\noindent Our key contributions can be summarized as follows:
\begin{itemize}
\item Introducing a Mahalanobis distance metric to GZSL, aiming to counteract the performance degradation due to projection bias. We also propose a novel loss function leveraging this distance metric for optimization.

\item A novel adaptation of the VAEGAN architecture featuring two discriminative modules, which is designed to address the GZSL challenge where training predominantly encounters seen class samples.

\item Robust experimental evidence demonstrating the superiority of our method over existing state-of-the-art techniques on four benchmark datasets.

\end{itemize}

\section{Related Work}
\label{sec:relatedwork}

Generalized zero-shot learning (GZSL) \cite{socher2013zero, chen2018zero} has garnered significant attention in the computer vision community as it holds promise in recognizing novel categories without explicit training samples. This section highlights the foundational works in GZSL and the methodologies employed to address the bias towards seen classes, culminating in the motivation for our approach.

\subsection{GZSL Foundations and Bias Challenges} Early GZSL approaches leveraged semantic embedding spaces derived from attributes or word vectors to bridge the gap between seen and unseen classes \cite{xian2018feature,changpinyo2016synthesized,fu2015transductive,xie2019attentive}. However, many of these methods were hindered by the challenge of bias towards seen classes, as the projection functions often relied heavily on seen data distributions \cite{Fu2015-transductive,ye2023rebalanced}. The bias problem was later formally analyzed \cite{pmlr-v97-zhang19}, revealing the projection's intrinsic limitations. In contrast to these methods, our work addresses the bias by incorporating a novel loss function that leverages the Mahalanobis distance, offering a more balanced treatment of seen and unseen classes.

\subsection{Projection Optimization} A significant portion of GZSL research has centred on optimizing the projection function. For instance, methods such as \cite{zhang2020towards,guan2019extreme} proposed complex mapping strategies to embed both seen and unseen samples into a shared semantic space. Techniques that integrate auxiliary information or adopt transductive settings to alleviate the bias have also been explored \cite{cheraghian2020transductive}.  Our method diverges by enhancing the projection optimization through a dual-branch VAEGAN architecture, which directly addresses the challenge of training predominantly with seen class samples.

\subsection{Distance Metric Learning} While optimizing the projection function remains a popular strategy, a few works have identified the importance of distance computation. During the training phase, some methods use Euclidean distance as a constraint to maintain the relationship between the generated visual features and the real semantic representation or use non-Euclidean embedding spaces \cite{mahdi2020-zeroshot} based on graph networks or manifold learning to maintain the relationship between data samples. However, the traditional Euclidean distance is still used in the inference stage to search for the nearest neighbour class of a given test sample. Our contribution uniquely focuses on the inference stage, where we apply the Mahalanobis distance to improve classification accuracy, distinguishing our approach from previous distance metric learning efforts.

\subsection{Generative Models in GZSL} 
Generative-models-based GZSL methods classify unseen samples using semantic representations. 
 Zero-Sample Learning Semantic Embeddings (SE-ZSL) \cite{frome2013devise} uses category embeddings, while Generative Zero-Sample Learning with Balanced Semantic Embeddings (LBSE-ZSL) \cite{xie2022leveraging} addresses category imbalance. Generative Zero Sample Learning using Visible and Invisible Semantic Relationships (LsrGAN) \cite{vyas2020leveraging} improves unseen category representation by leveraging class relationships. Guo et al. \cite{guo2023zero-shot} proposed an image-specific prompt learning (IPL) method, which produces a more precise adaptation for each cross-domain image pair, thereby enhancing the generator's flexibility. The integration of generative models like GANs and VAEs has shown promise in generating synthetic unseen class samples. For example, the VAEGAN model \cite{vaegan} fuses the strengths of VAEs and GANs to enhance generation quality. However, a unified architecture that robustly models seen and unseen data distributions remained an open challenge until our contribution. Our novel adaptation of the VAEGAN architecture introduces a mechanism for learning the Mahalanobis distance, setting our work apart by directly tackling the issue of data distribution modelling for both seen and unseen classes.

The bias towards seen classes in GZSL has been a consistent challenge, with most of the efforts focusing on refining the projection function. Our work diverges from this trend, emphasizing the significance of the distance metric during inference and extending the VAEGAN model to adaptively learn from both seen and unseen distributions. This positions our method distinctively in the GZSL landscape, as validated by our experimental results.

\section{Method}
\label{sec:method}
In this section, we first clarify the problem we aim to solve. Then, we introduce a modified VAEGAN framework. Next, we detail the integration of the Mahalanobis distance into the VAEGAN framework and propose a new loss function to facilitate learning the optimal Mahalanobis distance metric. Finally, we demonstrate how the Mahalanobis distance is utilized for classification during the inference stage.

\subsection{Problem Formulation}

The main goal of GZSL is to build a classifier based only on samples $\mathcal{X}^s$ of seen classes $\mathcal{C}^s$ that can simultaneously distinguish samples $\mathcal{X}^u$ from seen classes $\mathcal{C}^s$ and unseen classes $\mathcal{C}^u$, where unseen classes only appear in the test set, e.g. $\mathcal{C}^s \cap \mathcal{C}^u = \phi$. In addition to class labels, current existing methods fully use class-level semantic labels $\mathcal{S}$ (such as attributes or word2vec) to bridge the gap between seen and unseen classes. To this end, we define the training set as $\mathcal{D}^{tr} = \{I_i,\bm{s}_i,y_i)|I_i \in \mathcal{X}^s,\bm{s}_i \in \mathcal{S}, y_i \in \mathcal{C}^s\}$, where $I_i$ and $\bm{s}_i$ represent the image of the $i$-th sample and its semantic vector, respectively. Similarly, we can represent the test set by $\mathcal{D}^{te} = \{(I_i,\bm{s}_i,y_i)|I_i \in \mathcal{X}^s \cup \mathcal{X}^u,\bm{s}_i \in \mathcal{S}, y_i \in \mathcal{C}^s \cup \mathcal{C}^u \}$, where $I_i,\bm{y}_i$ either belong to the seen classes or belong to the unseen classes.
\begin{figure*}[!ht]
    \raggedright
    \includegraphics[width=1.0\textwidth]{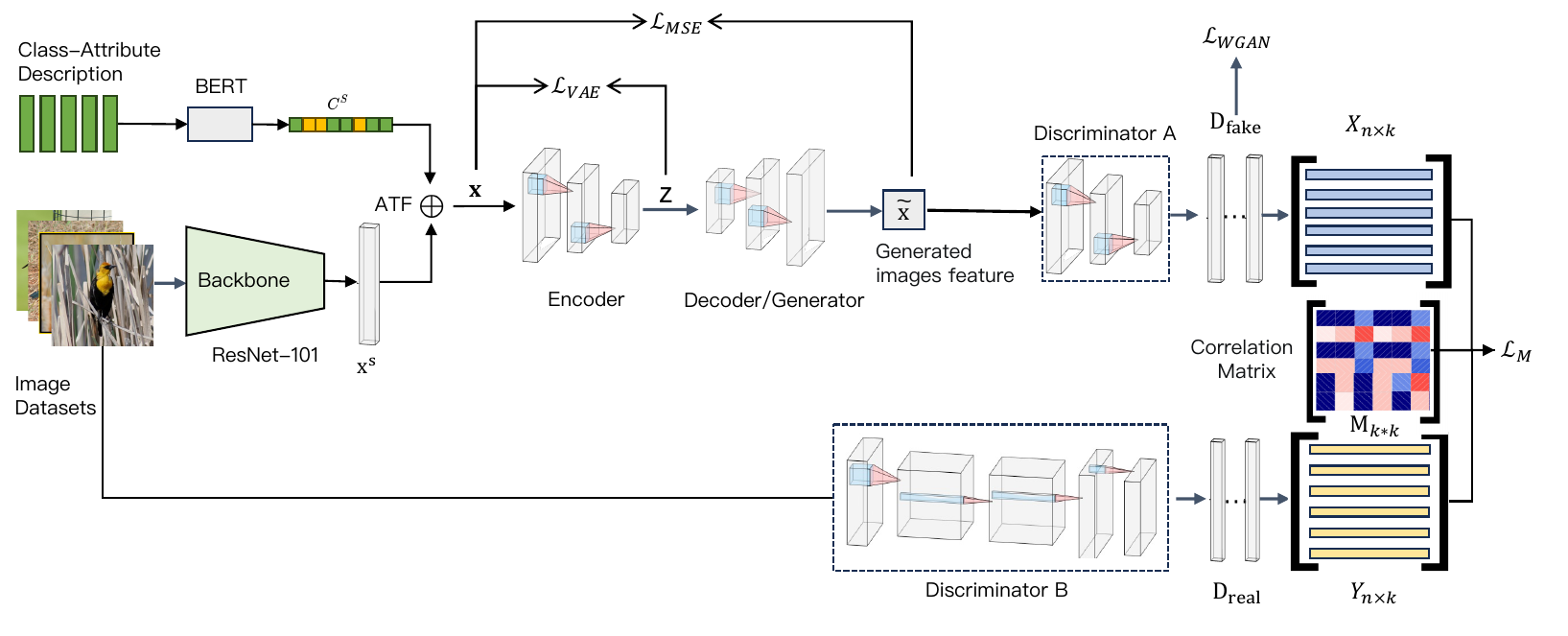}
    \caption{Framework of VAEGAN with Mahalanobis distance, featuring two branches: the upper branch generates images of unseen classes from seen classes using a generative network to simulate classification of unseen class images during the inference phase; the lower branch directly learns the projective representations of seen class images. The newly proposed Mahalanobis distance-based loss function aims to minimize the distance between projection outputs within the same branch while maximizing the distance between projections from different branches.}
    \label{fig:VAEGAN}
\end{figure*}

\subsection{Framework of VAEGAN}

VAEGAN \cite{xian2019f} combines the power of VAE models and WGAN models \cite{xian2018feature} to learn the data distribution of unlabeled samples by sharing the decoder in VAE and the generator in WGAN. We adopt this model to simulate scenarios where samples in the inference stage may come from unseen categories. During the training phase, the generated samples can be seen as coming from some fake unseen class, although these unseen classes are somewhat similar to the seen class.

\subsubsection{Feature Extraction} 

Under the framework of VAEGAN, image and text pairs $(I_i,\bm{s}_i)$ are processed through ResNet101 \cite{he2016deep}  or ViT-B \cite{dosovitskiy2020image}) and BERT \cite{liu2019roberta} to obtain their initial representation such as 

% replace the resnet-101 to vib-b/16

\eqn{}{
\bar{\bm{x}}_i = \textrm{ResNet101}(I_i) \in R^d, \quad \bar{\bm{s}}_i = \textrm{BERT}(\bm{s}_i) \in R^k.
}

We use the Affine Transformation Fusion (ATF)~\cite{tao2022df} to replace the common vector concatenation operation to better fuse the two multimodal information while keeping the dimensions unchanged, the schematic of ATF is shown in supplementary. We adopt two MLPs $\alpha(\cdot)$ and $\theta(\cdot)$ to predict the scale parameter and offset parameter of the affine transformation, respectively, as follows

\eqn{\label{eqn:aff-trans}}{
\bm{x}_i = \alpha(\bar{\bm{s}}_i)\bar{\bm{x}}_i + \theta(\bar{\bm{s}}_i) \in R^d,
}
where $\alpha(\cdot)$ will output a scalar, and $\theta(\cdot)$ will output a $d$-dimensional vector, 
 resulting in a fused representation of image $\bar{\bm{x}}_i$ and semantics $\bar{\bm{s}}_i$.  In the following description, unless otherwise specified, we will omit the subscript $i$.

% \begin{figure}[!ht]
%     \centering
%     \includegraphics[width=0.35\textwidth]{figs/af.png}
%     \caption{Affine Fusion Block Schematic \textcolor{blue}{SHREYANK: I think this can be moved to supplementary as it is borrowed work.}}
%     \label{fig1}
% \end{figure}

 \subsubsection{VAEGAN} A variational autoencoder (VAE) \cite{kingma2013auto} is a deep generative model capable of learning complex density model variables from latent data. Given a nonlinear generative model $p_{\phi}(\bm{x}|\bm{z})$, where $\bm{x}$ is the input of the network, the latent variable $\bm{z}$ comes from the prior distribution $p_0(\bm{z})$. The goal of a VAE is to approximate the posterior probability distribution of the latent variable $\bm{z}$ by maximizing the following variational lower bound through an inference network $q_{\tau}(\bm{z}|\bm{x})$
\eqn{}{
\mathcal{L}_{\phi,\tau} = \mathbb{E}_{q_{\tau}(\bm{z}|\bm{x})}[\log p_{\phi}(\bm{x}|\bm{z})] - \textrm{KL}(q_{\tau}(\bm{z}|\bm{x})||p_0(\bm{z})).
}
With the above consideration, we minimize the following the VAE loss \cite{xian2019f} with the input $\bm{x}$
\eqn{}{
\mathcal{L}_{\textrm{VAE}} = \textrm{KL}(q_{\tau}(\bm{z}|\bm{x})||p_0(\bm{z})) - \mathbb{E}_{q_{\tau}(\bm{z}|\bm{x})}[\log p_{\phi}(\bm{x}|\bm{z})],
}
where $q_{\tau}(\bm{z}|\bm{x})$ is an encoder $E(x)$, which encodes an input $\bm{x}$ to a latent variable $\bm{z}$, $p_{\phi}(\bm{x}|\bm{z})$ is a decoder, which reconstructs the input $\bm{x}$ from the latent $\bm{z}$ and the prior distribution $p_0(\bm{z})$ is assumed to be a standard normal distribution $\mathcal{N}(0,1)$.

It is worth noting that in VAEGAN, VAE's decoder $p_{\phi}(\bm{x}|\bm{z})$ and GAN's generator $G(\bm{z})$ share a network structure, so we use the discriminator $D_{A}(\bm{x})$ to distinguish real and fake samples, where the discriminator $D_{A}(\bm{x})$ will be optimized by minimizing the loss function
\eqna{
\mathcal{L}_{\textrm{WGAN}} & = & \mathbb{E}[D_A(\bm{x})] - \mathbb{E}[D_A(\tilde{\bm{x}})]  \nn \\
& - &  \lambda \mathbb{E}[\left(\|\nabla_{\tilde{\bm{x}}}D_A(\hat{\bm{x}})\|_2 - 1 \right)^2],
}
where $\tilde{\bm{x}} = G(\bm{z}) \sim p_{\phi}(\bm{x}|\bm{z})$, $\hat{\bm{x}} = \alpha \bm{x} + (1 - \alpha) \tilde{\bm{x}}$ and $\alpha \sim U(0,1)$.

Different from the literature \cite{xian2019f}, we effectively fuse the semantic information in the input and condition in VAEGAN through the learning of affine transformation (see Eqn. (\ref{eqn:aff-trans})). In addition, to ensure that the generated samples do not deviate too far from the real samples, we introduce the following MSE loss
\eqn{}{
\mathcal{L}_{\textrm{MSE}} = \mathbb{E}(\bm{x} - \tilde{\bm{x}})^2.
}

\subsection{Metric Learning with Stochastic Gradient Descent}
The root cause of projection bias is that in the projection space, the samples of the seen class and the samples of the unseen class are too close to each other (see Fig. \ref{fig:overview}). When the samples are classified according to the distance from the class description vector, the samples from the unseen class are likely to be classified into seen classes. To this end, we extend the traditional Euclidean distance metric to a general Mahalanobis distance metric so that under this distance metric, samples from unseen classes will be far away from the class vectors of seen classes, thereby improving the classification performance of GZSL.

Given two vectors $X$, $Y$ from projected space, we calculate the Mahalanobis distance between $X$ and $Y$ by the following formula
\eqn{\label{eqn:distance}}{
d_M^2(X,Y) = (X - Y)^T M (X - Y),
}
where $M$ is a positive definite matrix, which can clearly represent the correlation between the various components of the vector.

For the output $\tilde{\bm{x}}$ of VAEGAN and the image $I$, we simulate the projection output of unseen class samples and seen class samples respectively by two discriminators
\eqna{
X & = & D_{A}(\tilde{\bm{x}}) \in R^k, \label{eqn:discriminatora}\\
Y & = & D_{B}(I) \in R^k \label{eqn:discriminatorb}.
}
We stitch $N$ samples by row to get a $2N \times k$ matrix $\tilde{X}$ through the output $X$ and $Y$ of the two branches. In order to learn the optimal matrix $M$ under the framework of gradient descent, we represent $M$ in the following form
\eqn{}{
M = [\textrm{cov}(\tilde{X}) + \epsilon I ]^{+},
}
where $R^+$ represents the generalized inverse matrix of the matrix $R$. Note that $M$ is actually a function of network structure parameters, and it should be a symmetrical positive definite distance, thus ensuring that the distance (\ref{eqn:distance}) is an effective distance. 

Given a batch of samples, we propose a new loss function below so that the Mahalanobis distances of the projection outputs of the different branches are as far as possible, and the Mahalanobis distances of the projection outputs from the same branch are as close as possible
\eqn{}{
\mathcal{L}_{M} = - \log \sum_{i \neq j} \left(   d^2_{M}(X_i,Y_j) - d^2_{M}(X_i,X_j) \right),
}
where $X_i$ and $Y_j$ are calculated by Eqn. (\ref{eqn:discriminatora}) and (\ref{eqn:discriminatorb}), respectively.

Ultimately, our algorithm minimizes the following loss function via stochastic gradient descent (as shown in Alg. \ref{alg:flowchart})
\eqn{\label{eqn:total-loss}}{
\mathcal{L} = \mathcal{L}_{\textrm{WGAN}} + \lambda_{\textrm{VAE}} \mathcal{L}_{\textrm{VAE}} + \lambda_{\textrm{MSE}} \mathcal{L}_{\textrm{MSE}} + \lambda_{\textrm{M}} \mathcal{L}_{\textrm{M}},
}
where $\lambda_{\tm{VAE}}$, $\lambda_{\tm{MSE}}$ and $\lambda_{\tm{M}}$ are hyperparameters. Note that in optimizing the two discriminant models $D_A$ and $D_B$, our loss function is fundamentally different from one in f-VAEGAN-D2 \cite{xian2019f}: we optimize them by defining $\mathcal{L}_M$ loss, which is a simple minimization problem; but in f-VAEGAN-D2, their optimization is a classic min-max problem in GAN.

It is worth noting that since the matrix $M$ depends on the outputs of the two branches, $M$ is constantly updated during model iteration. We use $M^*$ obtained in the last iteration as the optimal metric for the inference phase.

\alg{\label{alg:flowchart}}{VAEGAN with Mahalanobis Metric}{
\st Input: A batch of images and class-attributes pairs $\langle I_i,\bm{s}_i,y_i\rangle$

\fr {$i = 1,2,\cdots$}
\st $\bar{\bm{x}}_i = \textrm{ResNet-101}(I_i)$ 
\st $\bar{\bm{s}}_i = \textrm{BERT}(\bm{s}_i)$
\st $\bm{x}_i = \alpha(\bar{\bm{s}}_i)\bar{\bm{x}}_i + \theta(\bar{\bm{s}}_i)$
\st Encode:  $\bm{z}_i = E(\bm{x}_i)$
\st Decode:  $\tilde{\bm{x}}_i = G(\bm{z}_i)$
\st Branch A: $X_i = D_A(\tilde{\bm{x}}_i)$
\st Branch B: $Y_i = D_B(I_i)$
\efr

\st $\tilde{X} = \textrm{cat}((X_1,X_2,\cdots,Y_1,Y_2,\cdots), \textrm{dim} = 0)$
\st $M = [\textrm{cov}(\tilde{X}) + \epsilon I]^{+}$
\st Compute the loss function $\mathcal{L}$ with Eqn. (\ref{eqn:total-loss})
}{
\tf{return} $\mathcal{L}$
}

When $M = I$, then the distance metric $d^2_{M}(X, Y)$ becomes an Euclidean distance. Since M is a positive definite distance, it has the following Cholesky decomposition form $M = L L^T$, and thus we have
\eqn{}{
d^2_{M}(X,Y) = \|L^T(X - Y)\|_2^2.
}
Unlike the projection bias problem in which the projection to vector $X$ is learned, our network optimizes the projection matrix of the difference $X - Y$ of any two vectors $X$ and $Y$. It is worth noting that the weight matrix in the Mahalanobis distance does not introduce additional parameters and only depends on the structural parameters of the network, which limits the complexity of the model and reduces the risk of model overfitting.

\subsection{Inference with Mahalanobis Metric}
In the inference stage (as shown in Alg. \ref{alg:inference}), for any image $I$, it and all class description text $s \in \mathcal{X}^s \cup \mathcal{X}^u$ are input into the upper branch of the network as multiple pairs, and the semantic representation $X$ of the class prototype is obtained through the discriminator $A$; at the same time, the image is directly passed through the lower branch of the network to obtain an embedded representation $Y$ of the image. Finally, according to the Mahalanobis distance between the image and the class prototype, the image will be classified into the nearest class.

\alg{\label{alg:inference}}{Inference with Mahalanobis Metric}{
\st Input: Any given image $I$
\st $\textrm{dist} = [ ]$
\st Branch B: $Y = D_B(I)$
\fr{$s \in \mathcal{X}^s \cup \mathcal{X}^u$}
\st $\bar{\bm{x}} = \textrm{ViT-B}(I)$ 
\st $\bar{\bm{s}} = \textrm{BERT}(\bm{s})$
\st $\bm{x} = \alpha(\bar{\bm{s}})\bar{\bm{x}} + \theta(\bar{\bm{s}})$
\st Encode:  $\bm{z} = E(\bm{x})$
\st Decode:  $\tilde{\bm{x}} = G(\bm{z})$
\st Branch A: $X = D_A(\tilde{\bm{x}})$
\st $d = d^2_{M^*}(X,Y)$
\st dist.append($d$)
\efr
\st $c = \argmax(\textrm{dist})$
}{\tf{return} $c$}

\section{Experiments}
\label{sec:experiments}

We first describe four popular public datasets and experimental implementation details. We then describe and compare the experimental implementation details with certain classical approaches. Finally, we conduct an ablation study to test the effectiveness of four important components of the work.

\subsection{Datasets and Evaluation Details}

We adopted four public datasets including Caltech-UCSD Birds-200-2011 (CUB) \cite{WahCUB_200_2011}, Animals with Attribute 1 (AWA1) \cite{lampert2009learning}, Animals with Attribute 2 (AWA2) \cite{xian2018zero}, SUN Database (SUN) 
 \cite{xiao2010sun}, and four other datasets. The CUB dataset contains 11,788 images of 200 species of birds, with about 60 images from each category. The AWA1 and AWA2 datasets each collect 50 different animal categories with about 40 to 60 images, and the AWA1 and AWA2 datasets have 30,475 and 37,322 images, respectively. The SUN dataset has images from 717 different scene categories, with about 200 to 500 images per category, totalling about 14,340. In dividing the dataset, we followed the conventional division method of GZSL datasets. 

 In the evaluation method, we used the harmonic mean $H$ to evaluate the recognition results on both visible and invisible class data simultaneously, which is often used to evaluate the classification performance of the GZSL task and is calculated as follows
\begin{equation}
H = \frac{2 \times U \times S}{U + S},
\end{equation}
where $U$ and $S$ denote the classification accuracy on unseen classes and seen class data, respectively.

\subsection{Implementation Details}

For basic visual features and visual extraction, we refer to VAEGAN \cite{xian2019f}. We use pre-trained ResNet-101~\cite{he2016deep} and Bert Tokenizer \cite{devlin2018bert} to extract the visual and semantic features of images and generate $2048$-dimensional visual feature vectors and $768$-dimensional semantic feature vectors, respectively. We also use  ViT-B \cite{dosovitskiy2020image} that generates $768$-dimensional visual feature vector for our generalization experiments. The dimension of the generated feature vector follows the initial hidden size of the model to ensure minimum feature loss \cite{dosovitskiy2020image, DBLP:journals/corr/abs-1810-04805}. These two vectors are fused together into a $32*3*384*384$ tensor for use in the encoding stage. Subsequently, we obtain a latent semantic representation of size $500$. The samples generated by the latent semantic representation pass through the discriminator A to obtain a vector of size $900$. Similarly, the original image also gets a $900$-dimensional vector through the discriminator B to facilitate our calculation of the Mahalanobis distance. The ADAM \cite{kingma2014adam} optimizer is used in our algorithm, where the learning rate is set to $10^{-3}$. We found that simply setting $\lambda_{\tm{VAE}}$, $\lambda_{\tm{MSE}}$ and $\lambda_{\tm{M}}$ to $1$, $1$, and $1$ gave the best results.

\subsection{Ablation Study}
\subsubsection{Mahalanobis/Euclidean Distance Metric}
\begin{table*}[!ht]
    \centering
    \caption{Performance of Our method with Euclidean Distance/Mahalanobis Distance}
    \resizebox{\textwidth}{!}{
    \begin{tabular}{|l|ccc|ccc|ccc|ccc|}
	    \hline \multirow{2}{*}{\textbf{Model}}  & \multicolumn{3}{|c|}{\textbf{CUB}} & \multicolumn{3}{c|}{\textbf{AWA1}} & \multicolumn{3}{c|}{\textbf{AWA2}} & \multicolumn{3}{c|}{\textbf{SUN}} \\
		\cline {2-13} & U & S & H & U & S & H & U & S & H & U & S & H \\
		\hline 
        $+$ Euclidean Distance & 16.9 & 41.6 & 24.0 & 18.1 & 46.3 & 26.1 & 19.4 & 43.2 & 26.8 & 9.7 & 18.4 & 12.4 \\
		\hline $+$ Mahalanobis Distance & \textbf{62.1} & \textbf{74.6} & \textbf{67.8}  & \textbf{67.2} & \textbf{76.3} & \textbf{71.5} & \textbf{64.9} & \textbf{79.1} & \textbf{71.3} & \textbf{45.7} & \textbf{49.8} & \textbf{47.7} \\
		\hline
	\end{tabular}}
    \label{tab:distance}
\end{table*}

The Mahalanobis distance plays a key role in our algorithm. To verify its effectiveness, we compare it with ordinary Euclidean distance. Since the Euclidean distance contains no parameters, we remove the $\mc{L}_{M}$ loss during training. The comparison results of using Euclidean distance and Mahalanobis distance on the four data sets are shown in Tab. \ref{tab:distance}.

From the experimental results in Tab. \ref{tab:distance}, the Euclidean distance measure performs very poorly under our model framework because our network uses two discriminator branches for learning Mahalanobis distance, which also shows the important role of Mahalanobis distance. We observe that the baseline architecture of VAEGAN is shown in Tab. \ref{tab:discriminator} also uses Euclidean distance, and the performance degradation of 4 data sets is not very obvious. %\tc{red}{However, it is worth noting our algorithm is simply a minimization optimization problem, rather than solving a minimax optimization problem in GAN.} 
At the same time, Mahalanobis distance takes into account the interaction between sample attribute features in the projection space. It is continuously optimized in the iterative process to better distinguish seen classes from unseen classes, alleviating the projection offset problem in the GZSL problem.

\subsubsection{Single-branch/Multi-branch Discriminator(s)}
In our work, we improve the structure of VAEGAN, in particular, we introduce another branch to better learn discriminative features between seen and unseen classes (see Tab. \ref{tab:discriminator}). On the other hand, we utilize two branches to define the loss $\mc{L}_M$ to learn the Mahalanobis distance metric. Note that Discriminator B cannot be used alone under our framework.

\begin{table*}[!ht]
    \centering
    \caption{Ablation study on our method with augmented discriminator}
    \resizebox{\textwidth}{!}{
    \begin{tabular}{|l|ccc|ccc|ccc|ccc|}
	    \hline
	    \multirow{2}{*}{\textbf{Model}} & \multicolumn{3}{c|}{\textbf{CUB}} & \multicolumn{3}{c|}{\textbf{AWA1}} & \multicolumn{3}{c|}{\textbf{AWA2}} & \multicolumn{3}{c|}{\textbf{SUN}} \\
	    \cline{2-13}
	    & U & S & H & U & S & H & U & S & H & U & S & H \\
	    \hline
	    VAEGAN with Discriminator A & 46.4 & 62.1 & 53.1 & 66.3 & 61.2 & 63.6 & 54.1 & 69.8 & 61.0 & 38.0 & 45.7 & 41.5 \\
	    \hline 
        $+$ Augmented Discriminator B & \textbf{62.1} & \textbf{74.6} & \textbf{67.8}  & \textbf{67.2} & \textbf{76.3} & \textbf{71.5} & \textbf{64.9} & \textbf{79.1} & \textbf{71.3} & \textbf{45.7} & \textbf{49.8} & \textbf{47.7} \\
	    \hline
    \end{tabular}}
    \label{tab:discriminator}
\end{table*}

\begin{table*}[!hbtp]
    \centering
    \caption{Performance comparison of our method under different information fusion}
    \resizebox{\textwidth}{!}{
    \begin{tabular}{|l|ccc|ccc|ccc|ccc|}
	    \hline \multirow{2}{*}{\textbf{Model}} & \multicolumn{3}{c|}{\textbf{CUB}} & \multicolumn{3}{c|}{\textbf{AWA1}} & \multicolumn{3}{c|}{\textbf{AWA2}} & \multicolumn{3}{c|}{\textbf{SUN}} \\
		\cline {2-13} & U & S & H & U & S & H & U & S & H & U & S & H \\
            \hline $+$ Concatenation operation & 56.0 & \textbf{81.6} & 66.4 & 61.0 & \textbf{82.3} & 70.1 & 61.1 & \textbf{80.4} & 69.4 & 33.7 & \textbf{52.2} & 41.0 \\
		\hline $+$ Affine transformation fusion (ATF) & \textbf{62.1} & 74.6 & \textbf{67.8} & \textbf{67.2} & 76.3 & \textbf{71.5} & \textbf{64.9} & 79.1 &  \textbf{71.3} & \textbf{45.7} & 49.8 & \textbf{47.7}\\
            \hline
	\end{tabular}
       }
    \label{tab:atf}
\end{table*}

According to the conducted experiments, our model greatly outperforms the baseline of VAEGAN in all categories except unseen categories in the AWA1 dataset (using the same ResNet features). This suggests that branch B of our model alleviates the projection bias/shift problem of samples to some extent and helps alleviate the problem related to semantic imbalance.

\subsubsection{Affine Transformation on Fusion/Concatenation}

The fusion of multiple modalities, such as images and texts, is usually fused by the concatenation operation. However, multiple modalities of the same sample are interrelated, and affine transformation fusion makes the change of text representation directly affect the mapping function to the image, realizing a closer information fusion between the two. In order to verify the effectiveness of the Affine Transformation Fusion (ATF) module under our model framework, we list the results of their comparison in Tab. \ref{tab:atf}. According to the experimental results, the affine transformation function shows a slight improvement over models that only capture image features and semantic features for all categories except the SUN dataset. As a result, the information fusion effect of this module is basically equivalent to the general concatenation operation.

% \subsubsection{Impact of Hyperparameters $\lambda_{\tm{VAE}}$, $\lambda_{\tm{MSE}}$ and $\lambda_{\tm{M}}$}
\subsubsection{Impact of Hyperparameters \texorpdfstring{$\lambda_{\text{VAE}}$}{lambda\_VAE}, \texorpdfstring{$\lambda_{\text{MSE}}$}{lambda\_MSE} and \texorpdfstring{$\lambda_{\text{M}}$}{lambda\_M}}

\begin{figure}[!ht]
    \centering
    \includegraphics[width=1.0\textwidth]{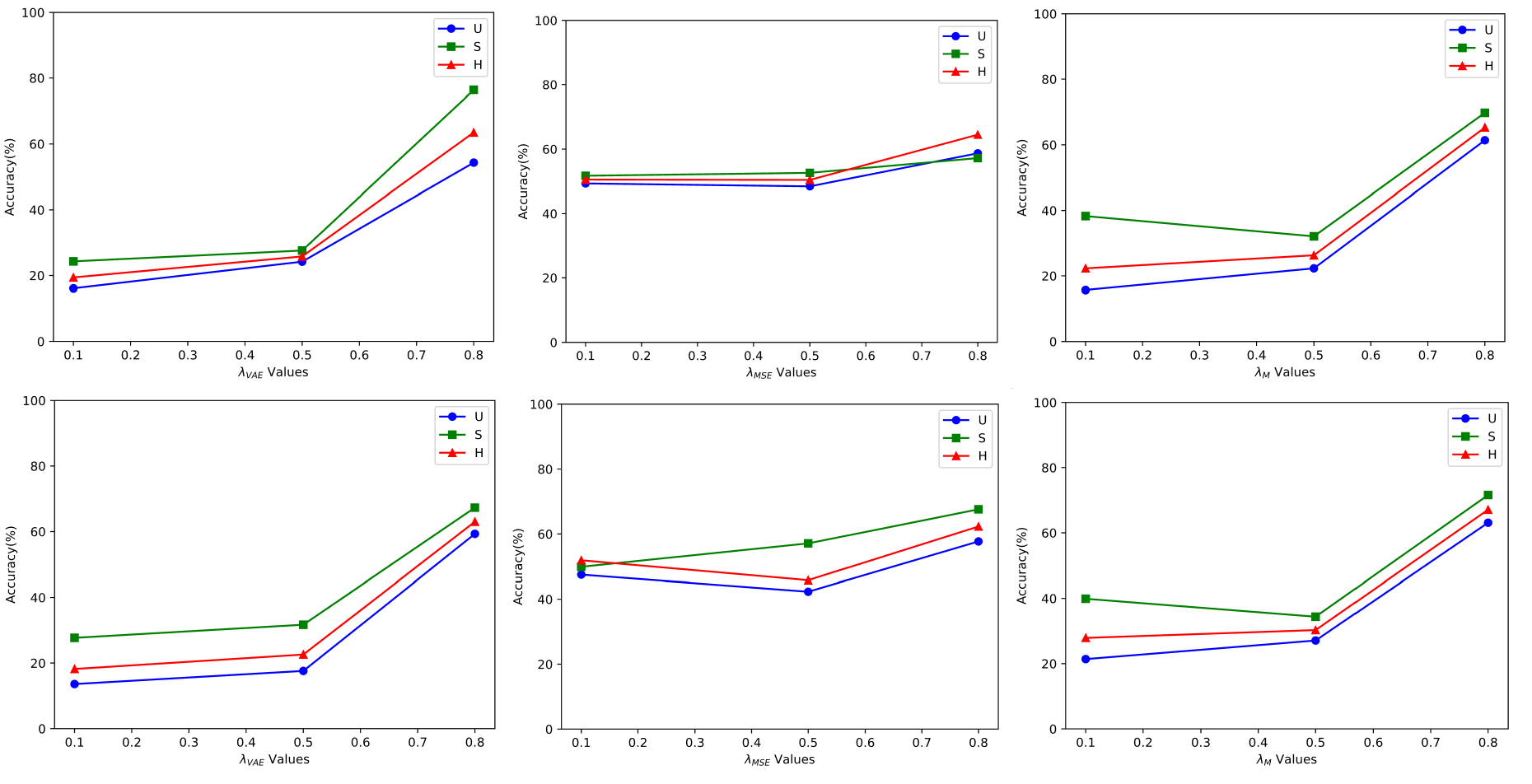}
    \caption{Performance of various $\lambda_{\tm{VAE}}$, $\lambda_{\tm{MSE}}$ and $\lambda_{\tm{M}}$ ratios on CUB dataset (above) and AWA2 dataset (below).}
    \label{fig:lambda_fig}
\end{figure}

We study the impact of the tradeoff parameters $\lambda_{\tm{VAE}}$, $\lambda_{\tm{MSE}}$ and $\lambda_{\tm{M}}$ on the performance of our algorithm. Fig. \ref{fig:lambda_fig} shows the recognition results of our method on seen classes and unseen classes under different parameter values, as well as their harmonic averages as shown in Tab. \ref{tab:losses}.
\begin{table*}[!ht]
    \caption{Comparison of the influence of different loss functions on model performance: $\mc{L}_{\tm{M}}$ based on Mahalanobis distance has the greatest impact, and the performance of the model drops the most when $\mc{L}_{\tm{M}}$ is removed (the comparative performance is marked in red and blue)}
    %\raggedbottom
    \centering
    \resizebox{1.0\textwidth}{!}{
    \begin{tabular}{|l|ccc|ccc|ccc|ccc|}
	    \hline \multirow{2}{*}{\textbf{Model}} & \multicolumn{3}{|c|}{\textbf{CUB}} & \multicolumn{3}{c|}{\textbf{AWA1}} & \multicolumn{3}{c|}{\textbf{AWA2}} & \multicolumn{3}{c|}{\textbf{SUN}} \\
		\cline {2-13} & U & S & H & U & S & H & U & S & H & U & S & H \\
        \hline 
            $\mathcal{L}_{\textrm{WGAN}}$ & 7.6 & 10.1 & 8.7 & 5.1 & 6.3 & 5.6 & 4.1 & 5.4 & 4.7 & 3.1 & 4.0 & 3.5\\
		\hline $\mathcal{L}_{\textrm{WGAN}}+\mathcal{L}_{\textrm{MSE}}+\mathcal{L}_{\textrm{M}} 
        (\lambda_{\tm{VAE}} = 0)$ & 15.8 & 25.7 & 19.6 & 15.7 & 27.1 & 19.9 & 14.4 & 26.1 & 18.6 & 12.3 & 17.9 & 14.6 \\
		\hline $\mathcal{L}_{\textrm{WGAN}}+\mathcal{L}_{\textrm{VAE}}+\mathcal{L}_{\textrm{M}} 
        (\lambda_{\tm{MSE}} = 0)$ & 43.7 & 49.9 & 46.6 & 38.8 & 52.3 & 44.5 & 42.3 & 55.6 & 48.0 & 38.8 & 41.5 & 40.1 \\
            \hline $\tc{blue}{\mathcal{L}_{\textrm{WGAN}}+\mathcal{L}_{\textrm{VAE}}+\mathcal{L}_{\textrm{MSE}}
         (\lambda_{\tm{M}} = 0)}$ & 8.8 & 27.3 & \tc{blue}{13.3} & 6.9 & 12.7 & \tc{blue}{8.9} & 7.6 & 14.7 & \tc{blue}{10.0} & 5.4 & 10.1 & \tc{blue}{7.0} \\
		\hline $\tc{red}{\mathcal{L}_{\textrm{WGAN}}+\mathcal{L}_{\textrm{VAE}}+\mathcal{L}_{\textrm{MSE}}+\mathcal{L}_{\textrm{M}}}$ & 62.1 & 74.6 & \tc{red}{67.8} & 67.2 & 76.3 & \tc{red}{71.5} & 64.9 & 79.1 & \tc{red}{71.3} & 45.7 & 49.8 & \tc{red}{47.7} \\
        \hline
	\end{tabular} }
    \label{tab:losses}
\end{table*}

\begin{table*}[!ht]
    \caption{Comparison of our method with other state-of-the-art methods on four datasets. Except LisGAN and ERPL, all baselines use the same 2048-D feature vectors of ResNet101 pretrained on ImageNet.} 
    \centering
    \resizebox{1.0\textwidth}{!}{%
    \begin{tabular}{|l|ccc|ccc|ccc|ccc|}
	    \hline \multirow{2}{*}{\textbf{Model (Year)}}  & \multicolumn{3}{|c|}{\textbf{CUB}} & \multicolumn{3}{c|}{\textbf{AWA1}} & \multicolumn{3}{c|}{\textbf{AWA2}} & \multicolumn{3}{c|}{\textbf{SUN}} \\
		\cline {2-13} & U & S & H & U & S & H & U & S & H & U & S & H \\
            \hline 
            \textbf{ERPL (2018)} & 43.7 & 37.2 & 40.2 & 66.4 & 50.4 & 57.4 & - & - & - & - & - & -  \\
            \textbf{LisGAN (2019)} & 46.5 & 57.9 & 51.6 & 52.6 & 76.3 & 62.3 & 54.3 & 68.5 & 60.6 & 42.9 & 37.8 & 40.2 \\
            \textbf{f-VAEGAN-D2 (2019)} & 48.4 & 60.1 & 53.6 & 62.9 & 63.3 & 63.5 & 57.6 & 70.6 & 63.5 & 45.1 & 38.0 & 41.3 \\
            \textbf{CADA-VAE (2019)} & 51.6 & 53.5 & 52.4 & 57.3 & 72.8 & 64.1 & 55.8 & 75 & 63.9 & 47.2 & 35.7 & 40.6 \\
		\textbf{CE-GZSL(2021)}& 54.2 & 67.2 & 61.4 & 65.3 & 73.4 & 69.1 & 63.1 & 78.6 & 70.0 & 48.8 & 38.6 & 43.1 \\
		\textbf{DCRGAN-TMM (2021)}& 40.6 & 54.1 & 46.4 & 32.2 & 65.0 & 43.1 & 51.6 & 63.8 & 57.1 & 40.5 & 34.6 & 37.3 \\
		\textbf{HSVA (2021)}& 52.2 & 59.7 & 55.7 & 61.1 & 75.2 & 67.4 & 57.4 & 81.1 & 67.3 & 48.6 & 39.0 & 43.3 \\  
         \textbf{BSeGN (2022)} & 55.3 &60.8 & 58.0 & - & - & - & 59.3 & 78.0 & 67.4 & 48.9 & 38.3 & 42.9 \\ 
            \textbf{DAZLE (2023)} & 56.7 & 59.6 & 58.1 & - & - & - & 60.3 & 75.7 & 67.1 & \textbf{52.3} & 24.3 & 33.2 \\
            \textbf{CMC-GAN (2023)} & 52.6 & 65.1 & 58.2 & 63.2 & 70.6 & 66.7 & - & - & - & 48.2 & 40.8 & 44.2 \\
            \textbf{DFTN (2023)} & 61.8 & 67.2 & 64.4 & 56.3 & \textbf{83.6} & 67.3 & 61.1 & 78.5 & 68.7 & - & -  & -  \\
            \textbf{CvDSF (2023)} & 53.7 & 60.0 & 56.9 & 64.5 & 71.4 & 67.8 & \textbf{65.6} & 70.4 & 67.9 & 49.2 & 38.0 & 42.9 \\
		\hline \textbf{Our model + ResNet101} & 57.1 & \textbf{81.6} & 67.2 & 62.9 & 83.1 & \textbf{71.6} & 62.2 & \textbf{82.3} & 70.9 & 39.6 & \textbf{52.7} & 45.9\\
            \textbf{Our model + ViT-B} & \textbf{62.1} & 74.6 & \textbf{67.8} & \textbf{67.2} & 76.3 & 71.5 & 64.9 & 79.1 & \textbf{71.3} & 45.7 & 49.8 & \textbf{47.7} \\
		\hline
	\end{tabular}}
    \label{tab:sota}
\end{table*}

As can be seen from Tab. \ref{tab:losses}, the model only using WGAN has weak discriminability. We remove different losses from Eqn. (\ref{eqn:total-loss}) to obtain the value of each loss's influence on the model, where the hyperparameter corresponding to the loss is set to $0$. It is obvious that the performance of each loss function is consistent on various data sets. For example, on the CUB data set, the loss $\mc{L}_{\tm{M}}$ has the greatest influence, which causes the harmonic mean $H$ of the model to drop from $67.8$ to $13.3$, followed by the loss $\mc{L}_{\tm{VAE}}$, and the loss $\mc{L}_{\tm{MSE}}$ has the smallest impact, that is $\mc{L}_{\tm{M}} > \mc{L}_{\tm{VAE}} > \mc{L}_{\tm{MSE}}$. Therefore, since the function of loss $\mc{L}_{\tm{M}}$ is to learn the Mahalanobis distance, it can be seen that the Mahalanobis distance plays a very important role. 

We assign values of $0.1$, $0.5$ and $0.8$ to one of $\lambda_{\tm{VAE}}$, $\lambda_{\tm{MSE}}$ and $\lambda_{\tm{M}}$ respectively, while keeping the other two parameters set to 1. As part of the ablation, we verified the impact of three parameter settings on algorithm performance on the data sets CUB and AWA2 and finally obtained 18 sets of experimental results, as shown in Fig. \ref{fig:lambda_fig}. On the one hand, for each loss, when we increase its weight, the performance of the algorithm is improved, especially the harmonic mean accuracy $H$. But $\mc{L}_{\tm{MSE}}$ is an exception. We need to set the parameters carefully to avoid some degree of overfitting. On the other hand, under the same weight, $\lambda_{\tm{M}}$ and $\lambda_{\tm{VAE}}$ have a greater impact on the algorithm. When the parameter increases from 0.1 to 0.8, the harmonic mean increases from about 20\% to more than 60\%. 

%%%%%%%%%%%%%%%%%%%%%%%%%%%%%%%%%%%%%%%%%%%%%%%%%%%%%%%%%%
% T-SNE feature vector figures with VAEGAN and our model %
%%%%%%%%%%%%%%%%%%%%%%%%%%%%%%%%%%%%%%%%%%%%%%%%%%%%%%%%%%

\subsection{Comparison with State-of-the-Art Methods}

% \textcolor{blue}{SHREYANK: \url{https://arxiv.org/pdf/2309.06987.pdf} More comprehensive comparison to SOTA in the link. Add year [2023] etc next to each paper in column 1 so that the reviewer knows papers have been cited all the way to the most recent ones.}

In this section, we select recent results of the state of the art in GZSL tasks, including LisGAN \cite{li2019leveraging}, f-VAEGAN-D2 \cite{xian2019f}, CADA-VAE \cite{schonfeld2019generalized}, CE-GZSL \cite{ han2021contrastive}, DCRGAN-TMM \cite{ye2021disentangling}, HSVA \cite{chen2021hsva}, DAZLE \cite{huynh2020fine}, DEM \cite{zhang2017learning}, CADA-VAE \cite{schonfeld2019generalized}, DAZLE \cite{huynh2020fine}, CMC-GAN \cite{yang2023semantics}, DFTN \cite{jia2023dual}, CvDSF \cite{zhai2023center} and BSeGN \cite{xie2022leveraging}, f-VAEGAN-D2 \cite{xian2019f}, ERPL \cite{guan2019extreme}, see Tab. \ref{tab:sota} for details. Some of the previous sota results are put into supplementary for reference, including ALE \cite{akata2015label}, f-CLSWGAN \cite{xian2018feature}.

Tab. \ref{tab:sota} lists the result comparison between our method and other classical methods. Specifically, in our method, the H-score can reach 67.2\% on the CUB dataset, 71.6\% on AWA1, 70.9\% on AWA2, and 45.9\% on SUN. Compared with the original VAEGAN model \cite{xie2022leveraging}: Our method improves the H-score of the model from 58.0\% to 67.2\% on the CUB dataset, from 67.4\% to 70.9\% on the AWA2, and finally, the SUN data set is increased from 42.9\% to 45.9\%. The above results show that our proposed method outperforms the state-of-the-art methods in all evaluated datasets and achieves significant improvements, especially in the $H$-score. We attribute the above results to three aspects: 1) Under the generative framework of VAEGAN, we simulate the output from seen samples and the output of unseen samples through two branches and make the samples of the two branches as far as possible through the loss function, at the same time, the samples of the same branch are as close as possible so that the seen class and the (pseudo) unseen class can be separated as much as possible; 2) Mahalanobis distance can help us correct wrong decisions when projection bias occurs, thereby improving classification performance on seen and unseen classes; 3) The method is backbone agnostic, replacing ResNet101 with ViT-B maintains high overall performance, showing the generalization of the approach. 

\section{Conclusions}

\label{sec:conclusions}
The biased projection is an important challenging problem in GZSL. In our work, we introduce the Mahalanobis distance into the VAEGAN framework. To this end, we use two branches to learn the samples of the seen class and the samples of the (pseudo) unseen class, respectively, and propose a new loss function such that the projected space is learned to be more discriminative for samples from unseen and seen classes. In particular, the weight matrix of the Mahalanobis distance does not introduce additional parameters, which limits the expressive ability of the model and avoids the possibility of further overfitting. Finally, our extensive experimental evaluation shows that our proposed method outperforms the state-of-the-art methods on four benchmark datasets. Our contribution has significant implications for advancing zero-shot learning and provides a promising avenue for future research in this area.

\bibliographystyle{unsrtnat}
\bibliography{references} 

\newpage

\appendix

\section{Affine Transformation Fusion Schematic}
\label{subsec:AFB}
\begin{figure}[!ht]
    \centering
    \includegraphics[width=0.7\textwidth]{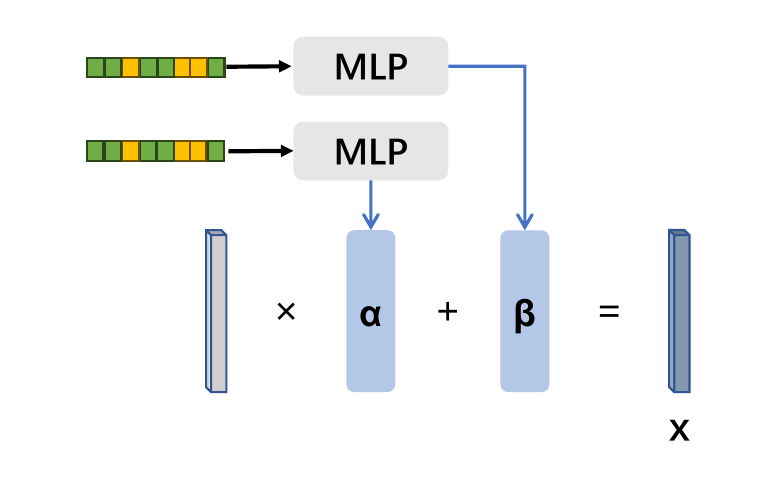}
    \caption{Affine Transformation Fusion schematic diagram}
    \label{fig1}
\end{figure}

For Affine Transformation Fusion, as shown in Fig. \ref{fig1}, we projected language-conditioned channel-wise scaling parameters $\alpha$ and shifting parameters $\beta$ from sentence vector $C^{s}$ from two MLPs (Multilayer Perceptron).

\begin{equation}
    \alpha = MLP_{1}(C^{s}), \quad\quad \beta = MLP_{2}(C^{s})
\end{equation}

for any given input feature from backbone $X^{s}$ we first
conduct the channel-wise scaling operation with the
scaling parameter $\alpha$, then apply the channel-wise shifting operation with the shifting parameter $\beta$,

\begin{equation}
    ATF(X^{s}, C^{s}) = \alpha X^{s}+ \beta = X
\end{equation}

where ATF denotes the Affine Transformation Fusion, $X^{s}$ is the image feature from backbone; $C^{s}$ is the sentence vector; $\alpha$ and $\beta$ are the scaling parameter and shifting parameter, respectively.

\section{More Results on Impact of \texorpdfstring{$\lambda_{\text{VAE}}$, $\lambda_{\text{MSE}}$ and $\lambda_{\text{M}}$}{lambda\_VAE, lambda\_MSE and lambda\_M} on Model Performance}

Tab. \ref{tab:lambda_sup} gives a more detailed data comparison of Figure 3. Each parameter takes
$\{0.1, 0.5, 0.8\}$, while keeping the other two parameter values as $1$, so that the impact of each parameter on the model can be quantitatively analyzed.
\begin{table*}[!ht]
    \caption{Quantitative results of the impact of parameters $\lambda_{\tm{VAE}}$,$\lambda_{\tm{MSE}}$ and $\lambda_{\tm{M}}$ on the model}
    \centering
    \begin{tabular}{|l|ccc|ccc|}
	    \hline  
             \multirow{2}{*}{$(\lambda_{\tm{VAE}},\lambda_{\tm{MSE}},\lambda_{\tm{M}})$} & \multicolumn{3}{|c|}{\textbf{CUB}} & \multicolumn{3}{c|}{\textbf{AWA2}}\\
		\cline {2-7} & U & S & H & U & S & H \\
		\hline $(0.1,1.0,1.0)$ & 16.1 & 24.3 & 19.4 & 13.6 & 27.7 & 18.2 \\
		$(0.5,1.0,1.0)$ & 24.2 & 27.6 & 25.8 & 17.6 & 31.7 & 22.6 \\
            $(0.8,1.0,1.0)$ & 54.3 & \tf{76.4} & 63.5 & 59.3 & 67.3 & 63.0 \\
            \hline $(1.0,0.1,1.0)$ & 49.3 & 51.7 & 50.5 & 47.5 & 57.1 & 51.9 \\
		$(1.0,0.5,1.0)$ & 48.4 & 52.6 & 50.4 & 42.3 & 49.9 & 45.8 \\
            $(1.0,0.8,1.0)$ & 58.7 & 57.2 & 64.5 & 57.7 & 67.6 & 62.3 \\
            \hline
            $(1.0,1.0,0.1)$ & 15.7 & 38.3 & 22.3 & 21.4 & 39.9 & 27.9 \\
            $(1.0,1.0,0.5)$ & 22.3 & 32.1 & 26.3 & 27.1 & 34.4 & 30.3 \\
            $(1.0,1.0,0.8)$ & 61.4 & 69.8 & 65.3 & 63.1 & 71.6 & 67.1 \\
            \hline
            $(1.0,1.0,1.0)$ & \tf{62.1} & 74.6 & \tf{67.8} & \tf{64.9} & \tf{79.1} & \tf{71.3} \\
            \hline
	\end{tabular}
    \label{tab:lambda_sup}
\end{table*}
% lambda 1,2,3 (0.1 0.5 0.8)

\begin{table*}[!ht]
    \caption{Comparison of our method with pre-2019 state-of-the-art methods on four datasets}
    \centering
    \resizebox{\textwidth}{!}{%
    \begin{tabular}{|l|ccc|ccc|ccc|ccc|}
	    \hline \multirow{2}{*}{\textbf{Model}} & \multicolumn{3}{|c|}{\textbf{CUB}} & \multicolumn{3}{c|}{\textbf{AWA1}} & \multicolumn{3}{c|}{\textbf{AWA2}} & \multicolumn{3}{c|}{\textbf{SUN}} \\
		\cline {2-13} & U & S & H & U & S & H & U & S & H & U & S & H \\
		\hline \textbf{ALE [2015]} & 23.7 & 62.8 & 34.4 & 16.8 & 76.1 & 27.5 & 14.0 & 81.8 & 23.9 & 21.8 & 33.1 & 26.3 \\
            \textbf{DEM [2017]}& 19.6 & 57.9 & 29.2 & 32.8 & 84.7 & 47.3 & 30.5 & 81.4 & 45.1 & 19.6 & 57.9 & 29.2 \\
		\textbf{f-CLSWGAN [2018]} & 31.73 & 64.34 & 42.50 & 61.41 & 59.63 & 60.51 & 29.85 & 76.60 & 42.96 & 42.6 & 36.6 & 39.4 \\
		\hline \textbf{Our model + ResNet50} & 57.1 & \textbf{81.6} & 67.2 & 62.9 & \textbf{83.1} & \textbf{71.6} & 62.2 & \textbf{82.3} & 70.9 & 39.6 & \textbf{52.7} & 45.9\\
            \textbf{Our model + ViT-B} & \textbf{62.1} & 74.6 & \textbf{67.8} & \textbf{67.2} & 76.3 & 71.5 & 64.9 & 79.1 & \textbf{71.3} & 45.7 & 49.8 & \textbf{47.7}\\
            \hline
	\end{tabular}}
    \label{tab:sota2}
\end{table*}

\section{Performance comparison of our method with pre-2019 methods}

Tab. \ref{tab:sota2} gives the performance comparison between our method and the method before 2019, as a supplement to the experimental results of the main text. It can be seen that the methods in 2019 mainly focus on the performance of seen classes (S columns), while the generalization on unseen (U columns) has been greatly improved. For example, on CUB, our method has greatly improved on invisible classes, where it reaches 62.1, compared with 31.73 of the f-CLSWGAN method.

\section{Performance on different dimensions of the feature vectors}

\begin{figure}[!ht]
	\centering
	\vspace{-15pt}
	\includegraphics[width=1.0\textwidth]{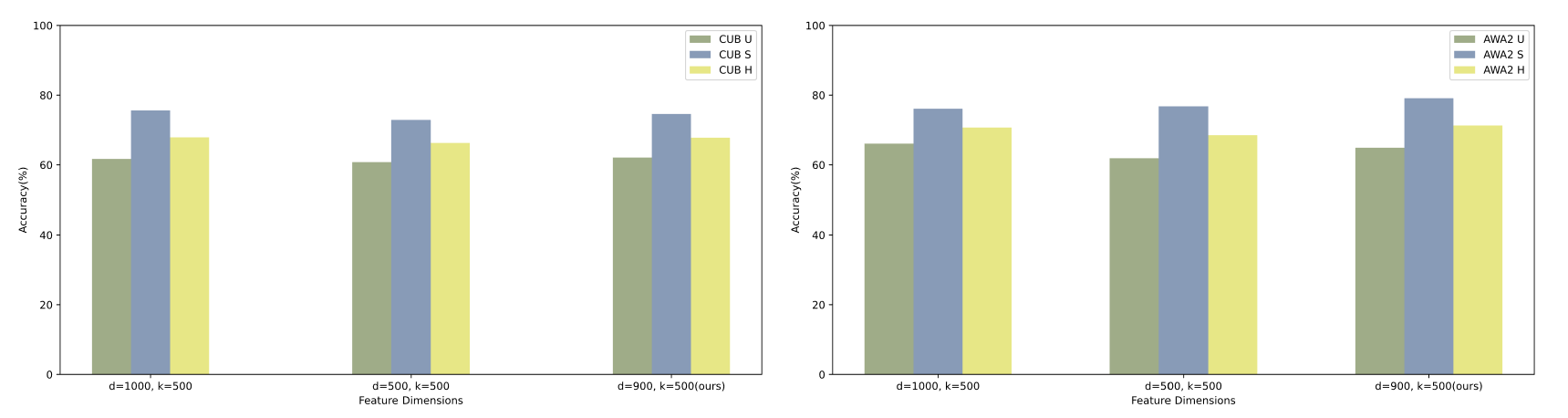}
	\caption{Impact of changes in the visual dimensions relative to the semantic dimensions of the dataset on model performance.}
	\label{fig:dim_fig}
	\vspace{-15pt}
\end{figure}

Regarding feature vector dimensions of image and semantic information, we take the hidden layer of each pre-trained model as the default output vector. At the same time, we also set the dimensions of the visual features to 500 and 1000 on the data sets CUB and AWA2 to explore the impact of the model parameters on the model performance, as shown in Fig. \ref{fig:dim_fig}.

It can be observed from Fig. \ref{fig:dim_fig} that the performance of the zero-shot experiments on the CUB and AWA2 datasets is less affected when the semantic dimension $k$ is fixed and the visualization dimension $d$ increases or decreases relative to the baseline value of $900$. On both datasets, increasing or decreasing the dimensionality of the visual features resulted in a slight improvement or decrease, but overall, the impact on model performance is not very large.

Tab. \ref{tab:dims} provides data support for Fig. \ref{fig:dim_fig}. We set up a group of experimental groups as our solution and used the other groups of dimensions as a comparison. We set the Latent semantic vector as $D_s$ and set the Visual vector as $D_v$. In the following table, we will adjust the dimensions of $D_s$ and $D_v$ to observe the performance change of the whole model and set the experimental group we use for comparison.

\begin{table*}[!ht]
    \caption{Performance comparison of the number of synthesised features on visual features}
    \centering
    \begin{tabular}{|l|ccc|ccc|ccc|ccc|}
	    \hline \multirow{2}{*}{\textbf{Model}} & \multicolumn{3}{|c|}{\textbf{CUB}} & \multicolumn{3}{c|}{\textbf{AWA2}} \\
		\cline {2-7} & U & S & H & U & S & H \\
		\hline  $D_v = 1000, D_s = 500$ & 61.7 & 75.6 & 67.9 & 66.1 & 76.1 & 70.7 \\
		\hline  $D_v = 500, D_s = 500$  & 60.8 & 72.9 & 66.3 & 61.9 & 76.8 & 68.5 \\
            \hline  $D_v = 900, D_s = 500$  & 62.1 & 74.6 & 67.8 & 64.9 & 79.1 & 71.3 \\
            \hline
	\end{tabular}
    \label{tab:dims}
\end{table*}

\begin{figure}[!ht]
    \centering
    \includegraphics[width=1.0\textwidth]{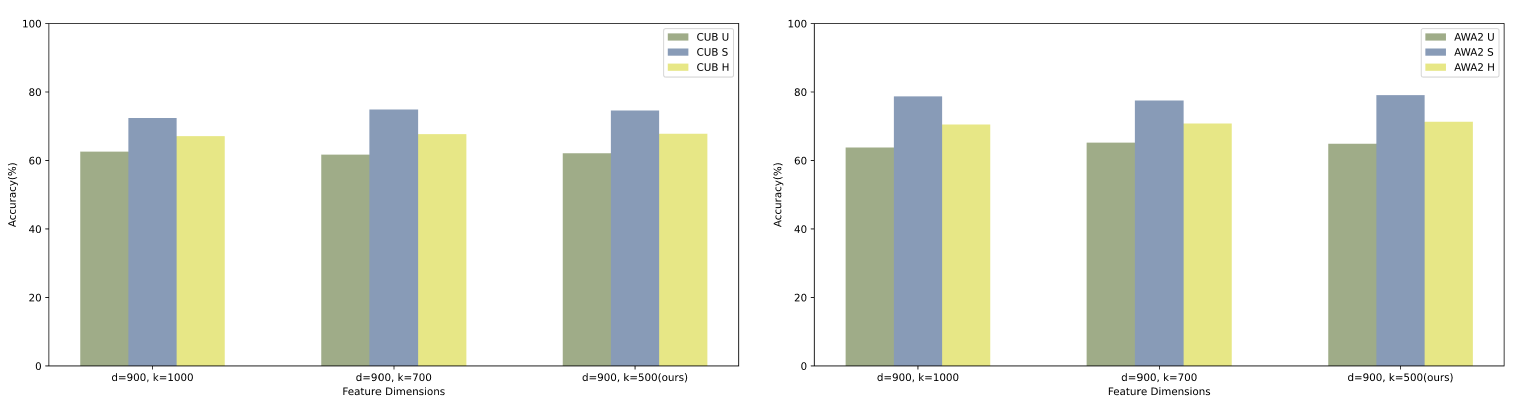}
    \caption{Impact of changes in the semantic dimensions relative to the visual dimensions of the dataset on model performance.}
    \label{fig:semantic-dim}
\end{figure}

Furthermore, we conducted comparative experiments on the semantic feature dimension. Specifically, we examined the effects of varying the size of the semantic feature dimension when the selected visual feature is set to 900. We tested three different sizes of semantic feature dimensions: \{300, 500, 700\}. Our analysis revealed that the size of the semantic feature dimension impacts the results when compared to our baseline. The results are shown in Tab. \ref{tab:dims-2} and Fig. \ref{fig:semantic-dim}.

\begin{table*}[!ht]
    \caption{Performance comparison of the number of synthesised features on semantic features}
    \centering
    \begin{tabular}{|l|ccc|ccc|ccc|ccc|}
	    \hline \multirow{2}{*}{\textbf{Model}} & \multicolumn{3}{|c|}{\textbf{CUB}} & \multicolumn{3}{c|}{\textbf{AWA2}} \\
		\cline {2-7} & U & S & H & U & S & H \\
            \hline  $D_v = 900, D_s = 1000$ & 62.6 & 72.4 & 67.1 & 63.8 & 78.7 & 70.5 \\
            \hline  $D_v = 900, D_s = 700$  & 61.7 & 74.9 & 67.7 & 65.2 & 77.5 & 70.8 \\
            \hline  $D_v = 900, D_s = 500$  & 62.1 & 74.6 & 67.8 & 64.9 & 79.1 & 71.3 \\
            \hline
	\end{tabular}
    \label{tab:dims-2}
\end{table*}

\section{Mahalanobis Metric and Euclidean Metric}
\fgr{\label{fig:ma-dist}}{Our method corrects the projection bias problem of the model through Mahalanobis distance: Mahalanobis distance-based inference correctly assigns the unknown sample from unseen classes to its true class Cat.}{0.45}{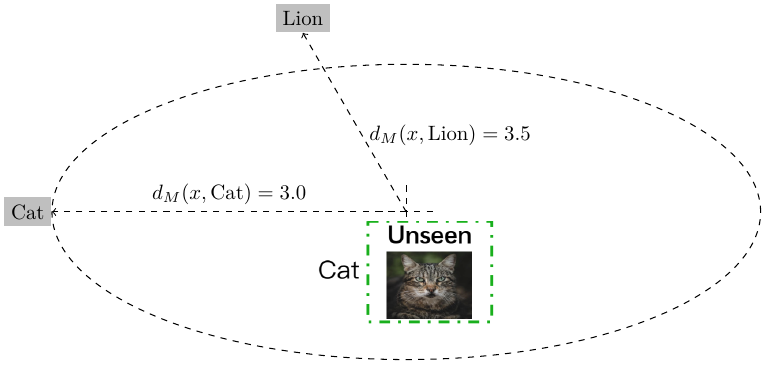}

Fig. \ref{fig:ma-dist} shows our motivation for using Mahalanobis distance instead of Euclidean distance in reasoning through a specific example of two similar classes Lion and Cat. First, we project unknown samples and semantic vectors into the same common space through a deep network model. According to the Euclidean distance, it can be seen that the unknown sample is closer to the seen class Lion, and then the unknown sample from the unseen class Cat will be mistakenly classified into the Lion class. However, according to the Mahalanobis distance, the unknown sample is closer to the class Cat. Note that the Mahalanobis distance from the dotted points of the ellipse to the centre of the ellipse is equal. Still, the category Lion is outside the dotted points of the ellipse, so the Mahalanobis distance between the category Lion and the centre of the ellipse is farther. Therefore, our method can alleviate the biased problem of projection learning in inference to some extent if our deep model learns a less accurate projection representation.

\section{T-SNE Visualization Results Comparison}

In this section, as shown in Fig. \ref{fig-tsne-1} and Fig. \ref{fig-tsne-2} below, we did four sets of T-SNE visualizations. The former group is a T-SNE visualization of the entire dataset features after extracting the data features of the dataset using ResNet50 and ViT-B on the CUB dataset, respectively. The latter two groups are T-SNE visualization after visual feature extraction of our dataset with VAEGAN and our model, respectively.

To reduce the dimensionality of our extracted features to 3 dimensions, we have set the dimension of the embedding space(n\_components) to 3. Additionally, we have set the random\_state to 42, the initialization method of the embedding space to PCA embedding, the perplexity to 50, and the number of iterations(n\_iter) of the optimization process to 2000.

\begin{figure*}[!ht]
    \centering
    \includegraphics[width=0.95\textwidth]{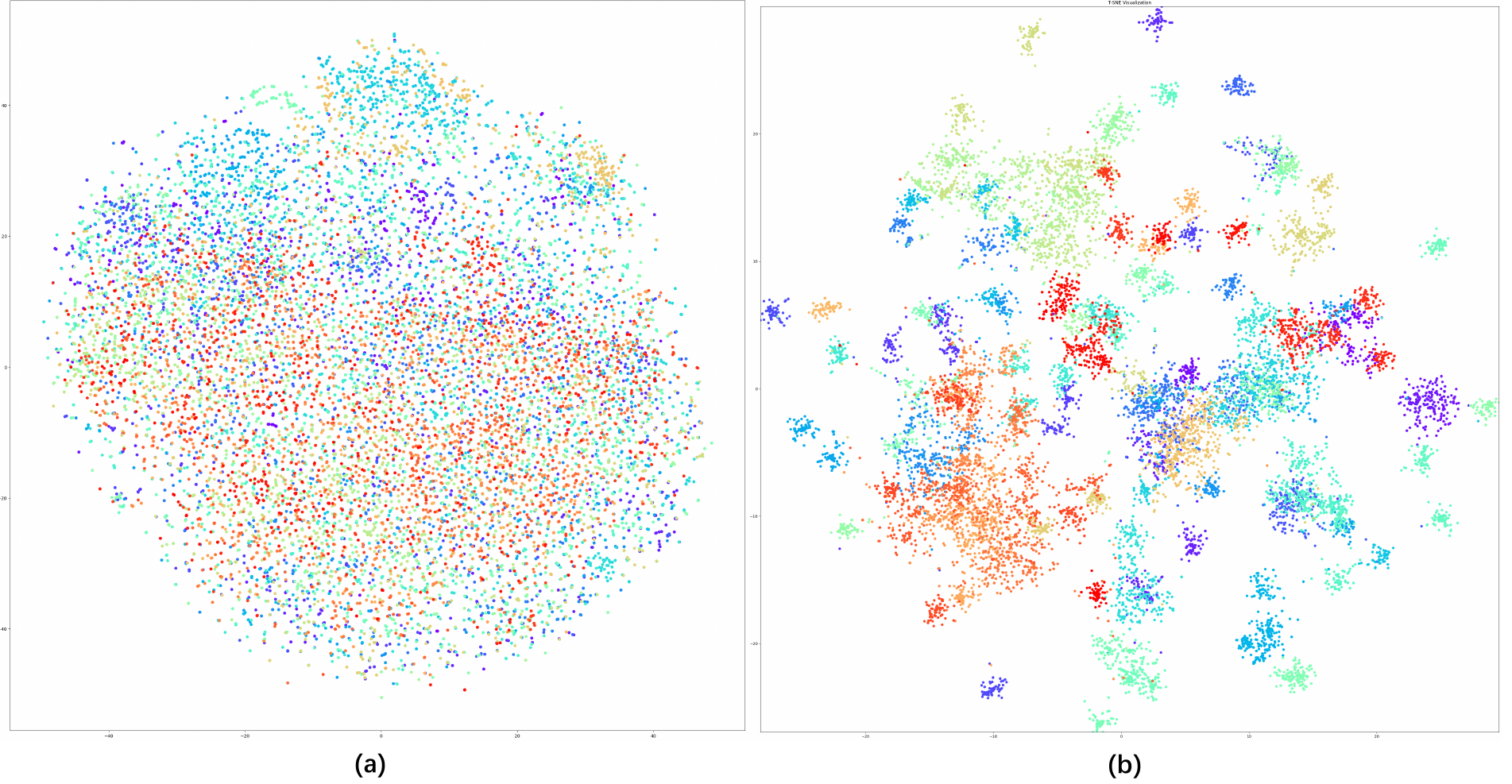}
    \caption{T-SNE visualization on CUB dataset. Fig(a). The visual feature extracted by ResNet50 backbone. Fig(b). The visual feature extracted by ViT-B backbone.}
    \label{fig-tsne-1}
\end{figure*}

\begin{figure*}[!ht]
    \centering
    \includegraphics[width=1.1\textwidth]{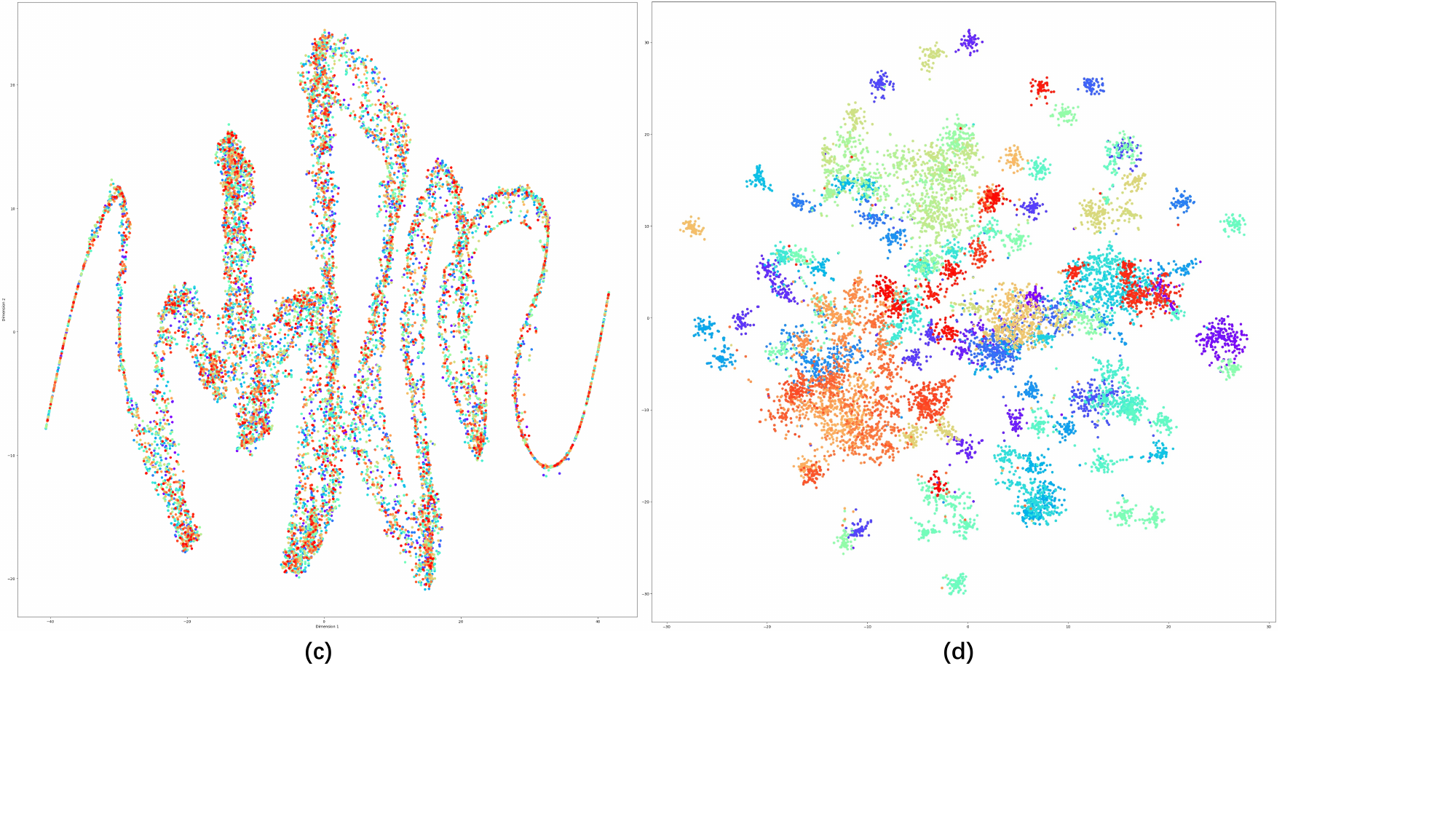}
    \vspace{-20pt}
    \caption{T-SNE visualization on CUB dataset. Fig(c). The visual feature extracted by VAEGAN. Fig(d). The visual feature extracted by our model.}
    \label{fig-tsne-2}
\end{figure*}

From the first set of results, ViT-B is effective in visual feature extraction thanks to ResNet50. After the former extracts the original visual features, the spatial distribution of visible and invisible classes is clearer, which lays the foundation for effective feature classification based on Mahalanobis distance in the next step. At the same time, it can be seen that ViT-B reduces the relative geometric relationship of entanglement between different classes, improves the performance and portability of the model, and plays a positive role in mapping from seen classes to unseen classes.

The second set of data is visualized by T-SNE to further explain why our method has a significant improvement over the previous one. Our method significantly improves the visual features of both seen and unseen classes, making the finger more capable of enhancing visual features and reducing classification complexity, thus increasing the recognition and transferability of the model.

\end{document}